\documentclass[10pt,twocolumn,letterpaper]{article}

\usepackage[pagenumbers]{cvpr} 
\usepackage{caption}
\usepackage{algorithm}
\usepackage{algpseudocode}
\usepackage{multirow}
\usepackage{enumitem}
\usepackage[dvipsnames]{xcolor}

\definecolor{cvprblue}{rgb}{0.21,0.49,0.74}
\usepackage[pagebackref=false,breaklinks,colorlinks,allcolors=cvprblue]{hyperref}

\def\paperID{4} 
\def\confName{CVPR}
\def\confYear{2026}
\newcommand{\formattedparagraph}[1]{\noindent \textbf{#1}}
\makeatletter
\def\blfootnote{\xdef\@thefnmark{}\@footnotetext}
\makeatother

\title{3D Gaussian Splatting for Efficient Retrospective Dynamic Scene Novel View Synthesis with a Standardized Benchmark}

\author{\quad Yunxiao Zhang$^{1, 2}$\quad Suryansh Kumar$^{1, 2, 3, 4, *}$\\
Visual and Spatial AI Lab${^1}$, VCCM Section  \\ College of PVFA${^2}$, Department of ECEN$^3$, Department of CSCE$^4$, \\ Texas A\&M University, College Station, Texas, USA
}

\begin{document}
\maketitle
\begin{abstract}
Retrospective novel view synthesis (NVS) of dynamic scenes is fundamental to applications such as sports. Recent dynamic 3D Gaussian Splatting (3DGS) approaches introduce temporally coupled formulations to enforce motion coherence across time. In this paper, we argue that, in a synchronized multi-view (MV) setting typical of sports, the dynamic scene at each time step is already strongly geometrically constrained. We posit that the availability of calibrated, synchronized viewpoints provides sufficient spatial consistency, and therefore, explicit temporal coupling, or complex multi-body constraints seems unnecessary for retrospective NVS. To this end, we propose an approach tailored for synchronized MV dynamic scene. By initializing the SfM-derived point cloud at the start time and propagating optimized Gaussians over time, we show that efficient retrospective NVS can be achieved without imposing a temporal deformation constraint. Complementing our methodological contribution, we introduce a Dynamic MV dataset framework built on Blender for reproducible NeRF and 3DGS research. The framework generates high-quality, synchronized camera rigs and exports training-ready datasets in standard formats, eliminating inconsistencies in coordinate conventions and data pipelines. Using the framework, we construct a dynamic benchmark suite and evaluate representative NeRF and 3DGS approaches under controlled conditions. Together, we show that, under a synchronized MV setup, efficient retrospective dynamic scene NVS can be achieved using 3DGS. At the same time, the dataset-generation framework enables reproducible and principled benchmarking of dynamic NVS methods. \blfootnote{*Corresponding Author: Suryansh Kumar}
\end{abstract}    

\section{Introduction}\label{sec:intro}
Retrospective novel view synthesis of dynamic scenes has become a central demand for modern visual media applications, including sports replay, performance analysis, immersive broadcasting, and virtual cinematography \cite{ZHANG2026104714}. In such scenarios, a dynamic event must not only be rendered from arbitrary viewpoints but also be revisited at any past moment with high quality image synthesis. This requires scene representations that are both temporally structured and scalable in compute time and memory footprint.

Recent advances in 3D scene representation, most notably Neural Radiance Fields (NeRF) \cite{mildenhall2020nerf} and 3D Gaussian Splatting (3DGS) \cite{kerbl20233d}, have substantially improved static and dynamic view synthesis. In particular, 3DGS has emerged as a compelling alternative to implicit radiance fields due to its real-time rendering capabilities and explicit geometric structure. Extensions such as 4DGS \cite{wu20244d} and space-time Gaussian \cite{li2024spacetime} variants have further incorporated temporal modeling to support dynamic scenes. However, two practical challenges remain underexplored.

\smallskip
\noindent
\textbf{(I)} The role of camera-object temporal motion coupling in dynamic scene modeling. It deserves closer examination in synchronized multi-view settings. In a general sports and visual-performance applications, cameras are mounted on a fixed platform and temporally synchronized, providing strong geometric constraints at every time step. Under such acquisition regimes, the scene at any given time $t$ is already well-posed under principle of multi-view rigidity in projective geometry \cite{hartley2003multiple}. This raises an important question: do we necessarily require complex temporally coupled deformation models to maintain coherence across frames, or can a carefully designed warm-start 3DGS pipeline suffice?

\smallskip
\noindent
\textbf{(II)} Dynamic novel view synthesis methods are often evaluated under inconsistent dataset conventions. Camera coordinate systems, temporal synchronization, training splits, and export formats frequently vary across implementations, complicating reproducibility and hindering quick and fair comparison between NeRF-based and 3DGS-based pipelines. As prior work shows, significant effort is often spent reconciling dataset formats and coordinate conventions before meaningful evaluation can even begin.

\begin{figure*}[t]
    \centering
\includegraphics[width=0.9\linewidth]{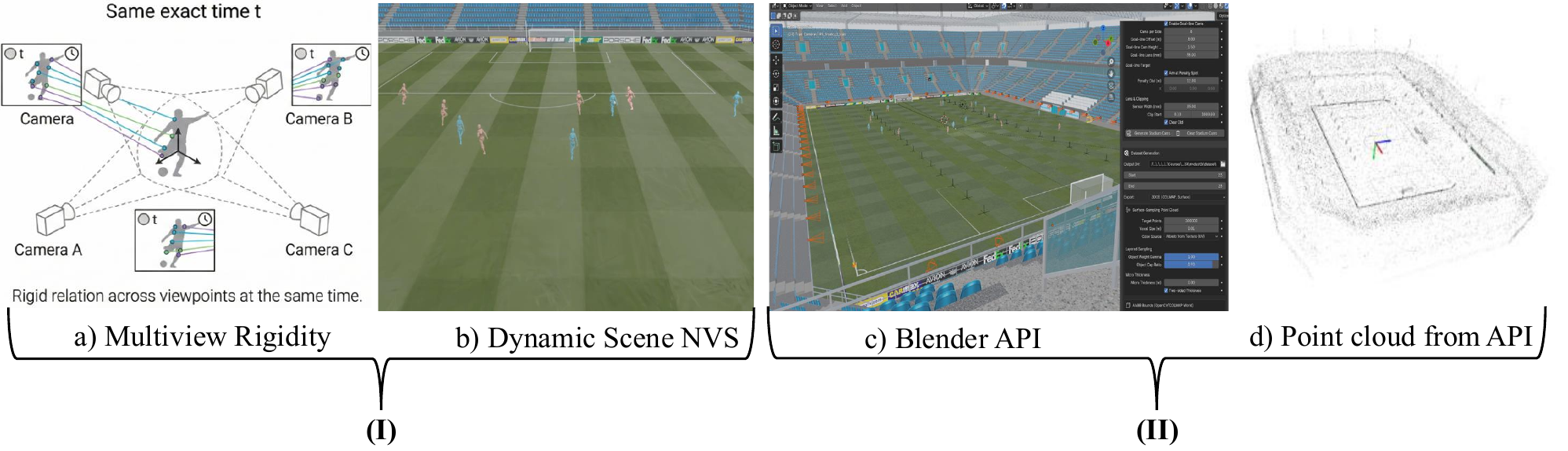}
\caption{\textbf{Contributions}: \textbf{(I)} a) The multi-body rigidity constraint holds in a dynamic scene under synchronized  multi-view camera setup for each time instance $t$, b) shows an example NVS results. \textbf{(II)} Proposed Blender API for generating multi-view dynamic scene dataset.}\label{fig:teaser_figure}
\end{figure*} 

In this work, we revisit dynamic scene modeling in a synchronized multi-view setup from the geometric perspective. We propose a warm-start 3DGS framework tailored for synchronized multi-view sequences. At the start, we initialize our approach using structure-from-motion (SfM) point cloud \cite{schonberger2016structure}, and subsequent frames are optimized via warm-start chaining, leveraging the fact that dynamic motion in sports is typically locally smooth between adjacent frames. Rather than introducing additional canonical-space deformations or temporal latent variables as done in \cite{wu20244d, luiten2024dynamic}, we show that, under well-calibrated, synchronized capture, geometry alone provides sufficient constraints for stable, memory efficient, and high-fidelity retrospective rendering.

Complementing the above methodological contribution, we introduce a Standardized Dynamic Multi-View Dataset Framework built on the popular open-source Blender  \cite{blender}. The framework automates the generation of parameterized synchronized camera rigs, enforces consistent coordinate conventions, and exports training-ready data compatible with both NeRF and 3DGS pipelines. It supports multi-frame dynamic sequences, camera-trajectory generation, and reproducible dataset splits. By eliminating dataset fragmentation and manual conversion steps, this framework substantially reduces experimental overhead and establishes a principled foundation for benchmarking dynamic neural rendering methods. Using this pipeline, we construct a dynamic benchmark suite targeting sports and visual-performance scenarios and systematically evaluate representative dynamic NeRF and 3DGS approaches under unified conditions. Our analysis highlights practical trade-offs between reconstruction fidelity, memory footprint, initialization sensitivity, and training scalability. In summary, this paper makes \textbf{two strong contributions}:
\smallskip
\begin{itemize}
    \item A geometrically grounded 3DGS formulation for efficient retrospective dynamic scene novel view synthesis (NVS) under synchronized multi-view capture.
    \item A standardized dynamic multi-view dataset and benchmarking framework for reproducible NeRF and 3DGS research, reducing experimental friction, manual effort and time while enabling principled comparative evaluation.
\end{itemize}
\smallskip
\noindent
Together, these contributions advance the methodological and experimental foundations of dynamic neural rendering, positioning 3DGS as a practical, reproducible solution for retrospective dynamic scene visualization. Figure \ref{fig:teaser_figure} provide the qualitative exposition to our contributions. {\href{https://github.com/JackZhang-SH/TimeArchival3DGS}{\textcolor{Brown}{Code}}, \href{https://github.com/JackZhang-SH/RealSynth_Studio}{\textcolor{Brown}{API}}, and \href{https://drive.google.com/drive/folders/13n8dcC3czco81el9sSp8ZaDHLI0m9HLw}{\textcolor{Brown}{Datasets}}} can be accessed from their respective links.

\section{Related Work}\label{sec:relatedwork}
NVS has long been a central problem in computer graphics \cite{seitz1996view, buehler2001unstructured}. Classical approaches laid the foundation for view interpolation and light-field rendering, while modern neural representations have substantially improved reconstruction quality and scalability. For comprehensive surveys, we refer readers to \cite{xiao2025neural, 10.1111:cgf.14505, chen2024survey}. In this work, we focus on recent developments to dynamic scene NVS using 3DGS techniques.

\smallskip
\formattedparagraph{\textit{(i)} Neural Scene Representations.} Early approaches parameterized scene content through volumes \cite{lombardi2019neural}, texture maps \cite{chen2020neural, thies2019deferred}, or neural point clouds \cite{aliev2020neural}. Not long ago, NeRF \cite{mildenhall2020nerf} approach to NVS transformed multi-view scene reconstruction by learning continuous volumetric scene representations from a set of well-posed image observations. NeRF marked a milestone by implicitly modeling scenes using differentiable volume rendering without explicit geometry, yielding a compact neural scene representation.

Subsequent work focused on improving NeRF's computational efficiency and expressiveness. Techniques such as Instant-NGP \cite{muller2022instant} used multi-resolution hash encodings for fast convergence, while methods such as TensoRF \cite{chen2022tensorf} achieved compact representations via decomposition. Other efforts in this direction improved sampling strategies \cite{attal2023hyperreel, neff2021donerf}, incorporated light-field representations or light information \cite{attal2022learning, feng2021signet, li2021neulf, suhail2022light, wang2022r2l, kaya2022neural, kaya2022uncertainty}, and explored multi-scale modeling \cite{barron2021mip, barron2022mip, jain2024learning, jain2022robustifying, jain2023enhanced}. Collectively, these works established implicit neural representations as a powerful framework for both static \cite{haghighi2023neural} and dynamic NVS \cite{ZHANG2026104714}.

\smallskip
\formattedparagraph{\textit{(ii)} 3D Gaussian Splatting.}
Lately, 3D Gaussian Splatting (3DGS) \cite{kerbl20233d} has emerged as a highly efficient alternative to purely implicit models such as NeRF. By representing scenes as anisotropic 3D Gaussians parameterized by position, covariance, opacity, and view-dependent color, 3DGS enables real-time rendering via differentiable rasterization. Its explicit geometric structure offers favorable convergence properties and high-fidelity image-synthesis quality.

\smallskip
\formattedparagraph{\textit{(iii)} Dynamic Scene Novel View Synthesis.}
Extending static scene representations for NVS to dynamic scenes is quite natural and has recently attracted significant interest. Early approaches leveraged multi-sphere imagery \cite{broxton2020immersive}, layered meshes \cite{bansal20204d}, and space-time encoding \cite{bemana2020x}. Neural Volumes \cite{lombardi2021mixture} and subsequent works modeled dynamic geometry through encoder-decoder architectures.

Meanwhile, NeRF-driven dynamic scenes approaches incorporate temporal conditioning \cite{li2022neural}, scene-flow \cite{li2021neural}, canonical deformation fields \cite{park2021nerfies, park2021hypernerf}, and factorized space-time representations \cite{shao2023tensor4d, fridovich2023k, cao2023hexplane}. These approaches use the well-known temporal smoothness and shared representations across frames. Likewise, the 3DGS extension to the dynamic scene has further expanded its applicability. Popular methods to this end include 4DGS \cite{wu20244d}, Dynamic 3D Gaussians \cite{luiten2024dynamic}, and space-time Gaussian \cite{yang2023real, yang2023deformable, lin2023gaussianflow}. More recent work includes MCMC-GS \cite{kheradmand20243d} that improves robustness with imperfect 3D point initialization. Compression methods such as QUEEN \cite{girish2024queen} address memory efficiency for dynamic Gaussian representations. While these recent developments demonstrate 3DGS's maturity and flexibility as a practical NVS framework, they are effective mostly for smoothly varying camera and shape motion.

On the contrary, this paper considers a synchronized multi-view camera setup. Under such a setup, unlike complex temporal coupling of dynamic subject(s), we propose a warm-start 3DGS strategy that leverages geometric continuity while maintaining per-frame optimization stability.


\smallskip
\formattedparagraph{\textit{(iv)} Reproducible Dynamic Scene Benchmarking.}
A practical challenge in dynamic scene rendering studies is the inconsistent conventions for datasets. Differences in coordinate systems, camera intrinsics, temporal synchronization, and export formats often complicate prompt and just comparison between NeRF-based and 3DGS-based methods. To this end, NeRF studio \cite{nerfstudio} provides an excellent platform, yet it is limited to rigid scenes. Datasets such as CMU Panoptic Studio \cite{joo2015panoptic} do provide high-quality multi-view captures, yet standardized pipelines for synthetic dynamic scene generation and benchmarking remain limited. To address these gaps, we introduce a standardized Dynamic Multi-View Dataset Framework built on the popular Blender \cite{blender}. The proposed API enforces consistent camera parameterization, synchronized temporal sampling, and export compatibility for both NeRF and 3DGS pipelines. By integrating scene design and training-ready data generation into a unified workflow, our API enables reproducible benchmarking of dynamic scene rendering methods.

\section{Method}\label{sec:method}
Unlike recent \textbf{dynamic 3DGS} methods \cite{wu20244d, luiten2024dynamic} that enforce temporal coherence via temporal coupling (e.g., deformation fields or canonical templates), we argue that in synchronized multi-view settings typical of sports, the scene at each time step is already strongly constrained by calibration and multi-view coverage. Consequently, memory efficient retrospective NVS can be achieved without imposing explicit temporal deformation constraints, by leveraging warm-start optimization in 3DGS pipeline across time.

\smallskip
\formattedparagraph{Problem Setup.}
We assume a synchronized multi-view capture setting with $N$ calibrated and time-synchronized cameras. At each discrete time instant $t \in \{1,\ldots,T\}$, the capture provides a set of RGB images
\begin{equation}\label{eq:It}
\mathcal{I}_t = \{ I_t^{(i)} \in \mathbb{R}^{H \times W \times 3} \}_{i=1}^{N},
\end{equation}
and known camera parameters $\pi_i = (K_i, R_i, \mathbf{t}_i)$, where $K_i$ is the intrinsic matrix and $(R_i,\mathbf{t}_i)$ map world coordinates to camera coordinates. The problem we aim to solve is given a query time $t$ and a novel camera viewpoint $\pi^\star = (K^\star, R^\star, \mathbf{t}^\star)$, we aim to synthesize $\hat{I}_t^\star$ corresponding to the dynamic scene state at time $t$ from viewpoint $\pi^\star$, while enabling random access in time.

\smallskip
\formattedparagraph{\textit{(i)} Time-Indexed Fixed-Size 3D Gaussian Representation.} We represent the scene at each time $t$ as a fixed-size set of $K$ anisotropic 3D Gaussians:

\begin{equation}
\mathcal{G}_t = \{ g_{t,k} \}_{k=1}^{K}.
\end{equation}
Each Gaussian $g_{t,k}$ is parameterized by its mean $\boldsymbol{\mu}_{t,k} \in \mathbb{R}^3$, covariance $\Sigma_{t,k} \in \mathbb{R}^{3 \times 3}$, opacity $\alpha_{t,k} \in (0,1)$, and view-dependent appearance represented by spherical harmonics (SH) coefficients $\mathbf{b}_{t,k}$. We factorize the covariance as
\begin{equation}
\Sigma_{t,k} = R(\mathbf{q}_{t,k}) \, \mathrm{diag}(\mathbf{s}_{t,k}^2)\, R(\mathbf{q}_{t,k})^\top,
\end{equation}
where $\mathbf{q}_{t,k}$ is a unit quaternion and $\mathbf{s}_{t,k} \in \mathbb{R}^3_{+}$ are per-axis scales. We efficiently store the full dynamic event in the set

\begin{equation}
\mathcal{A} = \{ \mathcal{G}_1,\ldots,\mathcal{G}_T \},
\end{equation}
enabling retrospective access by loading $\mathcal{G}_t$ for any $t$. Differently from standard 3DGS training \cite{kerbl20233d}, we \textbf{disable densification} so that $|\mathcal{G}_t| = K$ for all $t$ to yields predictable memory footprint, archive size (low-memory footprint), and rendering throughput over long sequences.

\smallskip
\formattedparagraph{\textit{(ii)} Differentiable Rasterization via Gaussian Splatting.}
Given a camera $i$, we transform Gaussian centers to camera coordinates:
\begin{equation}
\tilde{\boldsymbol{\mu}}^{(i)}_{t,k} = R_i \boldsymbol{\mu}_{t,k} + \mathbf{t}_i.
\end{equation}
Let $\Pi(\cdot)$ denote perspective projection. The 2D mean on the image plane is
\begin{equation}
\mathbf{u}^{(i)}_{t,k} = \Pi\!\left(K_i \tilde{\boldsymbol{\mu}}^{(i)}_{t,k}\right) \in \mathbb{R}^2.
\end{equation}
To obtain a screen-space elliptical footprint, we compute the Jacobian $J$ of the projection at $\tilde{\boldsymbol{\mu}}^{(i)}_{t,k}$ and form the 2D covariance approximation
\begin{equation}
\Sigma^{(i)}_{t,k,2D} = J \, \left(R_i \Sigma_{t,k} R_i^T\right)\, J^T.
\end{equation}
The Gaussian kernel at pixel $\mathbf{p}\in\mathbb{R}^2$ is
\begin{equation}
G^{(i)}_{t,k}(\mathbf{p}) =
\exp\!\left(
-\tfrac{1}{2}(\mathbf{p}-\mathbf{u}^{(i)}_{t,k})^T
(\Sigma^{(i)}_{t,k,2D})^{-1}
(\mathbf{p}-\mathbf{u}^{(i)}_{t,k})
\right),
\end{equation}
and $a^{(i)}_{t,k}(\mathbf{p}) = \alpha_{t,k} G^{(i)}_{t,k}(\mathbf{p})$ as its per-pixel alpha contribution. By letting $\mathbf{d}^{(i)}_{t,k}$ be the normalized viewing direction from the camera center to $\boldsymbol{\mu}_{t,k}$ and using the SH basis functions $\{Y_m(\cdot)\}$, the {view-dependent color} is implemented as: 
\begin{equation}
\mathbf{c}^{(i)}_{t,k} = \sum_m \mathbf{b}_{t,k,m}\, Y_m(\mathbf{d}^{(i)}_{t,k}) \in \mathbb{R}^3.
\end{equation}

\noindent
For alpha compositing the Gaussians are depth-sorted and composited front-to-back:
\begin{align}
\hat{I}^{(i)}_t(\mathbf{p})
&=
\sum_{k=1}^{K}
T^{(i)}_{t,k-1}(\mathbf{p})\; a^{(i)}_{t,k}(\mathbf{p})~\mathbf{c}^{(i)}_{t,k},
\\
T^{(i)}_{t,k}(\mathbf{p})
&=
\prod_{j=1}^{k}\left(1-a^{(i)}_{t,j}(\mathbf{p})\right).
\end{align}
This formulation for rasterization is differentiable with respect to Gaussian parameters, enabling gradient-based optimization from images.

\smallskip
\formattedparagraph{\textit{(iii)} Point-cloud initialization.} We use structure from motion (SfM) derived point cloud only at the start time $t_0$ (typically $t_0=1$). Given $\{I_{t_0}^{(i)}\}_{i=1}^{N}$ and the calibrated camera setup, we compute a point set
\begin{equation}
\mathcal{P}_{t_0}=\{\mathbf{p}_k \in \mathbb{R}^3\}_{k=1}^{K},
\end{equation}
by selecting/sampling $K$ points from the reconstructed cloud (to match the fixed Gaussian budget). We initialize:
\begin{equation}
\boldsymbol{\mu}_{t_0,k}\!\leftarrow\! \mathbf{p}_k,\quad
\mathbf{s}_{t_0,k}\!\leftarrow\! s_0\mathbf{1},\quad
\mathbf{q}_{t_0,k}\!\leftarrow\! \mathbf{q}_0,\quad
\alpha_{t_0,k}\!\leftarrow\! \alpha_0.
\end{equation}
We initialize appearance coefficients $\mathbf{b}_{t_0,k}$ by projecting $\mathbf{p}_k$ into views where it is visible and aggregating observed colors (e.g., via median). The SH average value of the lighting function over a sphere is set to the aggregated RGB value, and higher-order coefficients are initialized to zero.

\begin{figure}[t]
\centering
\includegraphics[width=1.0\linewidth]{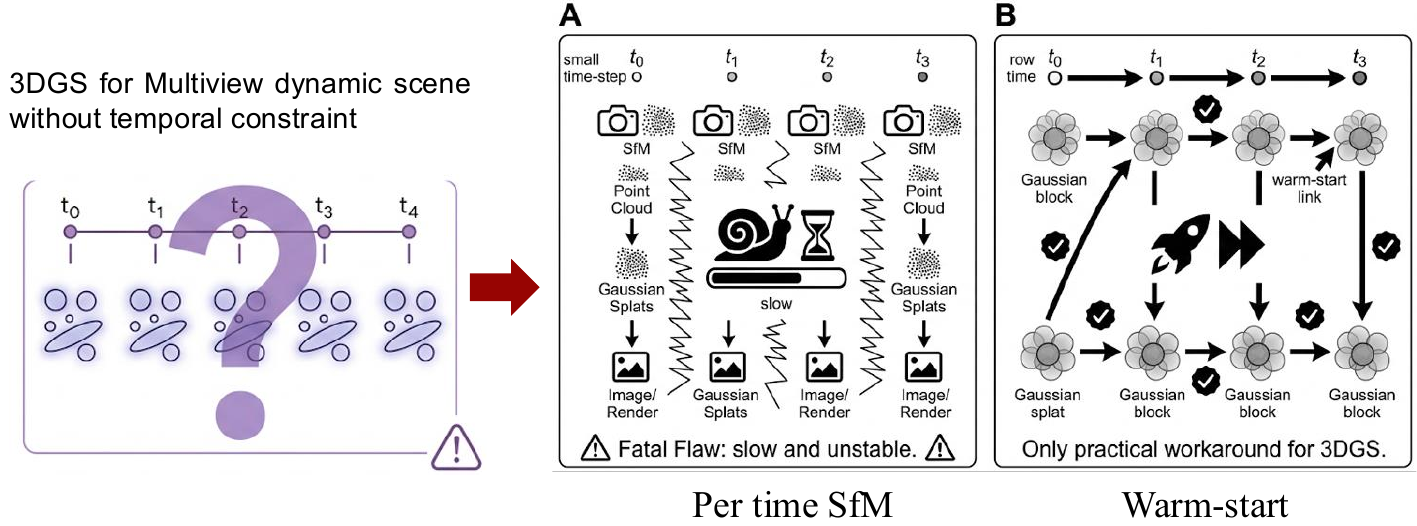}
\caption{For complex highly dynamic scene using constraint proposed in 4DGS \cite{wu20244d} often break down. On the contrary initializing 3DGS with SfM point cloud for each time instance and envoking rigidity constraint will make the entire pipeline slow. We propose Warm-start (\textbf{Right}) training across time, where each frame start from a neighboring frame converged parameter.}\label{fig:warm-start-motivation}
\end{figure} 

\smallskip
\formattedparagraph{\textit{(iii)} Per-Time-Step Optimization Without Explicit Temporal Deformation.} For each time $t$, we optimize the Gaussian parameters

\begin{equation}
\Theta_t = \{ \boldsymbol{\mu}_{t,k}, \Sigma_{t,k}, \alpha_{t,k}, \mathbf{b}_{t,k} \}_{k=1}^{K}
\end{equation}
by minimizing a multi-view reconstruction objective:
\begin{equation}
\mathcal{L}_t(\Theta_t)
=
\sum_{i=1}^{N}
\mathcal{L}_{\mathrm{img}}\!\left(\hat{I}^{(i)}_t(\Theta_t), ~I^{(i)}_t\right)
+
\lambda_{\mathrm{reg}} \,\mathcal{R}(\Theta_t).
\label{eq:total_loss}
\end{equation}
We instantiate $\mathcal{L}_{\mathrm{img}}$ using robust photometric losses (e.g., $\ell_1$ / Charbonnier) optionally combined with a structural term such as SSIM. The regularizer $\mathcal{R}$ is lightweight and aims to avoid degenerate solutions, e.g.,
\begin{equation}
\mathcal{R}(\Theta_t) =
\sum_{k=1}^{K}
\left(
\|\mathbf{s}_{t,k}\|_2^2 + \alpha_{t,k}^2 + \|\mathbf{b}_{t,k}\|_2^2
\right),
\end{equation}
together with clamping $\mathbf{s}_{t,k}$ to a valid interval. Importantly, we do not introduce any explicit temporal deformation model or canonical-time constraint into Eq.~\eqref{eq:total_loss}. In synchronized multi-view capture, spatial consistency is enforced per time by calibration and multi-view supervision. 

\smallskip
\formattedparagraph{\textit{(iv)} Warm-Start (Warm-Chain) Training Across Time.} Training an independent 3DGS model from scratch for each time $t$ is costly and introduces unnecessary variability. We therefore adopt \textbf{warm-chain initialization}, where each frame starts from a neighboring frame's converged parameters. Let $\Theta_t^\star$ denote the optimized parameters at time $t$. For a forward warm chain, we initialize
\begin{equation}
\Theta_t^{(0)} \leftarrow \Theta_{t-1}^\star \quad \text{for } t=t_0+1,\ldots,T,
\end{equation}
and optimize Eq.~\eqref{eq:total_loss} for a fixed number of steps.
Equivalently, we can process frames in reverse order (backward warm chain):
\begin{equation}
\Theta_t^{(0)} \leftarrow \Theta_{t+1}^\star \quad \text{for } t=T-1,\ldots,t_0.
\end{equation}
In both cases, warm start is used strictly for initialization and does not impose an explicit temporal constraint. Note. we disable densification/splitting such that the number of Gaussians remains fixed for efficient archival (low-memory footprint) of the event, i.e., 
\begin{equation}
|\mathcal{G}_t| = K \quad \forall ~t.
\end{equation}
This design ensures stable compute cost per frame and constant archive size, which is desirable for long sports events. Fig. \ref{fig:warm-start-motivation} show a visualization to our Warm-start strategy.

\smallskip
\formattedparagraph{\textit{(v)} Retrospective Novel-View Synthesis and Archival.} After optimization, we archive $\Theta_t^\star$ (or equivalently $\mathcal{G}_t$) for each $t$, forming $\mathcal{A}$. Retrospective NVS is performed by loading $\Theta_t^\star$ for a desired time and rendering from an arbitrary virtual camera $\pi^\star$ using the splatting as detailed above. This supports random access in time and viewpoint. Algorithm  \ref{alg:warmstart_3dgs} provide the pseudo-code of our implementation.

Our proposed warm-start, densification-free formulation for dynamic scene NVS yields:
1) stable memory and archive size (fixed $K$ per frame), 2) stable rendering throughput (constant Gaussian budget), and 3) efficient per-frame training (warm initialization reduces optimization effort), while maintaining rendering quality at an acceptable level under strong synchronized multi-view constraints.

\begin{algorithm}[t]
\caption{Efficient Retrospective Dynamic 3DGS}
\label{alg:warmstart_3dgs}
\begin{algorithmic}[1]
\Require Synchronized images $\{\mathcal{I}_t\}_{t=1}^{T}$, camera calibrations $\{\pi_i\}_{i=1}^{N}$, Gaussian budget $K$
\State Choose start time $t_0$ (typically $t_0=1$)
\State \textbf{SfM init at $t_0$:} compute point cloud $\mathcal{P}_{t_0}$ and sample $K$ points
\State Initialize $\Theta_{t_0}^{(0)}$ from $\mathcal{P}_{t_0}$; optimize to obtain $\Theta_{t_0}^\star$
\For{$t = t_0+1$ to $T$} \Comment{forward warm chain (or reverse the loop)}
    \State Warm start: $\Theta_t^{(0)} \leftarrow \Theta_{t-1}^\star$ \Comment{no densification; $K$ fixed}
    \State Optimize Eq.~\eqref{eq:total_loss} to obtain $\Theta_t^\star$
    \State Archive $\Theta_t^\star$
\EndFor
\State \textbf{Render:} for any query $(t,\pi^\star)$, load $\Theta_t^\star$ and render $\hat{I}_t^\star$
\end{algorithmic}

\end{algorithm}

\begin{figure*}[t]
    \centering
\includegraphics[width=\linewidth]{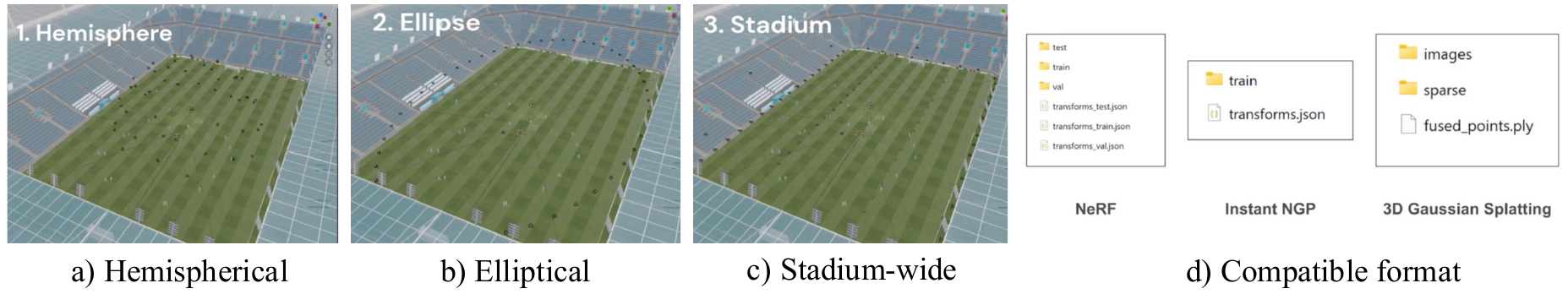}
\caption{Showing different camera configurations (in black frustum) as well as file format than can be generated using the proposed API.}\label{fig:blender-api}
\end{figure*} 

\section{Blender API for Dynamic NVS}
We introduce a Blender-based dataset generation API that operationalizes the paper's second contribution. It is a reproducible, format-consistent pipeline for constructing synchronized multi-view, multi-frame benchmarks. The API provides parameterized rig synthesis for common capture geometries, including hemisphere, full sphere, ellipse ring, and a stadium sports layout (ring + tiled cameras with optional goal-line viewpoints). Furthermore, it can export temporally aligned image sequences together with calibrated camera metadata (see Fig.\ref{fig:blender-api}). To eliminate the frequent ambiguity in coordinate conventions across NeRF and 3DGS codebases, Blender is used strictly for authoring, while export/import is normalized to an OpenCV/COLMAP-style world-to-camera convention via explicit axis-conversion matrices. The intrinsics follow a top-left image-origin convention to match standard photogrammetry tool-chains. 

The developed exporter supports multiple downstream pipelines, including Instant-NGP \cite{muller2022instant}, NeRF Synthetic \cite{mildenhall2020nerf}, TACV \cite{zhang2026timearchivalcameravirtualizationsports}, and COLMAP \cite{schonberger2016structure} poses-only. For 3DGS workflows that depend on geometry or depth, we additionally provide three depth and point-cloud modes: \textbf{(1) COLMAP + surface}, which samples a sparse point cloud from the visible mesh surface and exports it alongside COLMAP poses, \textbf{(2) COLMAP + depth}, which back-projects rendered per-view depth maps into 3D and fuses them into a point cloud with COLMAP poses, and \textbf{(3) depth-only}, which exports per-camera depth maps (e.g., \texttt{.npy} and \texttt{.pt}) without fusion. Finally, to ensure protocol-level reproducibility, train/validation/test splits are deterministically assigned by camera name/suffix, making the split effectively invariant over time when the same camera set is reused---a property that is essential for controlled temporal benchmarking of dynamic NVS methods.

\section{Experiment and Results}

\formattedparagraph{Implementation Details.} 
We implemented our approach using PyTorch 2.5.1 and conducted all experiments on NVIDIA GPUs with CUDA 11.8. All experiments are performed on an NVIDIA A40 GPU (50 GB RAM). 

\subsection{Dataset for Evaluation}
In this section, we briefly summarize the synthetic dynamic scene datasets used for evaluation. For consistency,  we keep dataset construction, camera setup, and train/validation/test split set same as proposed recently in \cite{zhang2026timearchivalcameravirtualizationsports}.

\subsubsection{Synthetic Multiview Dynamic Scene Dataset.}
We imported three synthetic datasets proposed in \cite{zhang2026timearchivalcameravirtualizationsports} using our benchmarking API in Blender 4.0 for evaluation: \textbf{\textit{(i)}} Dancing-Walking-Standing (\textbf{D-WS}), \textbf{\textit{(ii)}} Soccer Penalty Kick (\textbf{S-PK}), and \textbf{\textit{(iii)}} Soccer Multiplayer (\textbf{S-MP}). The 3D models and hemispherical experimental setup used to create these datasets is same as \cite{zhang2026timearchivalcameravirtualizationsports}. Under the synchronized multiview camera setup, for each discrete time instance we acquire $\mathcal{I}_t$ (refer to Eq.~\eqref{eq:It}).

\begin{table*}[t]
\centering
\resizebox{\textwidth}{!}{
\begin{tabular}{l|c|c|c|c|c|c|c|c|c|c}
\toprule
{Method$\rightarrow$} & 
{{D-3DGS \cite{luiten2024dynamic}}} & 
{{4DGS \cite{wu20244d}}} & 
{{ST-GS \cite{li2024spacetime}}} &
{{D-NeRF \cite{pumarola2021d}}} &
{{T-4D \cite{shao2023tensor4d}}} & 
{{HP \cite{cao2023hexplane}}} & 
{{KP \cite{fridovich2023k}}} & 
{{S-RF \cite{li2022streaming}}} &
TACV \cite{zhang2026timearchivalcameravirtualizationsports} &
\textbf{TA-3DGS} \\
\midrule
{\textbf{D-WS}} & & & & & & & & & & \\
\hline
PSNR    & 18.45 & 28.17 & 20.03 & 6.44  & 16.55 & 15.82 & 16.40 & 18.87 & 34.28 & \textbf{42.50} \\
LPIPS   & 0.1396 & 0.0800 & 0.1120 & 0.5726 & 0.2165 & 0.2470 & 0.2327 & 0.2054 & 0.0275 & \textbf{0.0110} \\
Memory  & 2.0MB & 21.0MB & 3.2MB & 512k  & 280MB & 68MB & 419MB & 2.2GB & 3.1GB & 4.94GB \\
Train Time & 1.41h & 0.83h & 0.46h & 4.25h & 10.02h & 0.96h & 1.02h & 0.65h & 5.25h & 2.76h \\
Iterations & 130K & 17K & 30K & 20K & 200K & 25K & 30K & 80K & 19K & \textbf{8k} \\
\midrule
{\textbf{S-PK}} & & & & & & & & & & \\
\hline
PSNR    & 26.45 & 26.25 & 25.99 & 10.64 & 25.86 & 22.45 & 21.30 & 22.28 & 33.81 & \textbf{44.59} \\
LPIPS   & 0.0719 & 0.0450 & 0.0778 & 0.4070 & 0.0588 & 0.1567 & 0.1866 & 0.1792 & 0.0282 & \textbf{0.0023} \\
Memory  & 2.0MB & 13.9MB & 0.9MB & 197K & 280MB & 68MB & 419MB & 1.8 GB & 5.2GB & 4.25GB \\
Train Time & 1.48h & 3.02h & 0.47h & 7.74h & 9.12h & 0.75h & 0.94h & 0.55h & 8.90h & 4.68h \\
Iterations & 130K & 17K & 30K & 20K & 200K & 25K & 30K & 80K & 16.5K & \textbf{8k} \\
\midrule
{\textbf{S-MP}} & & & & & & & & & & \\
\hline
PSNR    & 26.43 & 26.20 & 25.92 & 6.15 & 25.82 & 20.42 & 19.34 & 24.98 & 31.85 & \textbf{43.84} \\
LPIPS   & 0.0872 & 0.061 & 0.104 & 0.5330 & 0.0881 & 0.2104 & 0.2115 & 0.0906 & 0.0392 & \textbf{0.0023} \\
Memory  & 2.0MB & 19MB & 0.9MB & 228K & 280MB & 68M & 419MB & 1.8GB & 4.0GB & 3.81GB \\
Train Time & 1.46h & 2.94h & 0.71h & 4.21h & 10.1h & 0.83h & 1.03h & 0.58h & 6.28h & 3.44h \\
Iterations & 130K & 17K & 30K & 20K & 200K & 25K & 30K & 80K & 16.5K & \textbf{8k} \\
\bottomrule
\end{tabular}
}
\caption{Comparison of our results with popular dynamic NeRF and dynamic 3DGS approaches on three synthetic datasets. By using the warm-start up strategy with the multiview rigidity in a synchronized camera setup, we achieve best PSNR and LPIPS score. The notable part is that our approach despite using the fixed number of Gaussians, it demonstrate high-quality results for Dynamic NVS.}
\label{tab:comparison_result_with_both_3dgs_and_nerf_approaches}
\end{table*}

\smallskip
\formattedparagraph{\textit{(i)} Dancing-Walking-Standing (D-WS).}  
This dataset contains three dynamic subjects exhibiting diverse motion patterns, including dancing, walking, and standing. A total of 65 time instances are rendered, each from 100 calibrated camera viewpoints. We designate the $0^\textrm{th}$, $30^\textrm{th}$, $60^\textrm{th}$, and $90^\textrm{th}$ cameras as test views, while the $1^\textrm{st}$ camera is used for validation. This yields 95 training views, 1 validation view, and 4 test views per time step, resulting in 6,175 training images ($65 \times 95$), 65 validation images ($65 \times 1$), and 260 test images ($65 \times 4$). The split remains fixed across all time steps to ensure reproducible benchmarking.

\smallskip
\formattedparagraph{\textit{(ii)} Soccer Penalty Kick (S-PK).} 
This dataset consists of two dynamic subjects simulating a soccer penalty kick scenario. We render 109 time instances, each from 60 camera viewpoints. The $21^\textrm{st}$, $37^\textrm{th}$, $40^\textrm{th}$, and $56^\textrm{th}$ cameras are used for testing, while the $0^\textrm{th}$ camera is reserved for validation. This results in 55 training views, 1 validation view, and 4 test views per frame, corresponding to 5,995 training images ($109 \times 55$), 109 validation images ($109 \times 1$), and 436 test images ($109 \times 4$). All cameras are temporally synchronized to ensure consistent multiview capture.

\smallskip
\formattedparagraph{\textit{(iii)} Soccer Multiplayer (S-MP).} 
This dataset captures three dynamic subjects performing coordinated soccer actions. It contains 83 time instances rendered from 60 camera viewpoints. Among these, 55 viewpoints are used for training, 1 for validation, and 4 for testing. Specifically, the $21^\textrm{st}$, $37^\textrm{th}$, $40^\textrm{th}$, and $56^\textrm{th}$ cameras are used as test views, and the $0^\textrm{th}$ camera is used for validation. This produces 4,565 training images ($83 \times 55$), 83 validation images ($83 \times 1$), and 332 test images ($83 \times 4$). As with the other datasets, synchronized cameras are used throughout for evaluation.

\smallskip
\formattedparagraph{\textit{(iv)} Soccer field in different weather condition.}
We further propose three soccer field datasets under different weather conditions, namely \textit{Cloudy SoccerField}, \textit{Snow SoccerField}, and \textit{Sunny SoccerField}, generated using our Blender plugin. It contains 120 time instances rendered from 60 camera viewpoints. Among these, 58 viewpoints are used for training and 2 for testing. This produces 6960 training images ($58 \times 120$), and 240 test images ($2 \times 120$). All images are rendered at $1920 \times 1080$ resolution.

\begin{table*}[t]
\centering
\resizebox{\textwidth}{!}{
\begin{tabular}{lcccccccc}
\toprule
Dataset & Total Train Img & Test Img & Iter. for \textbf{A} & Iter. for \textbf{B} & PSNR $\uparrow$ & LPIPS $\downarrow$ & Merging Time & Total Memory (\textbf{A} + \textbf{B} filtered)\\
\midrule
Cloudy SoccerField & 6960  & 2 & 30000 & 5200 & 28.15 & 0.370 & 0.92h & 886MB \\
Snow SoccerField   & 6960  & 2 & 30000 & 5200 & 35.00 & 0.270 & 0.66h & 478MB \\
Sunny SoccerField  & 6960  & 2 & 30000 & 5200 & 28.51 & 0.337 & 1.46h & 888MB \\
\bottomrule
\end{tabular}
}
\caption{TA-3DGS results on the new soccer field datasets using H100 with $\textbf{A/B}$ separate reconstruction.}
\label{tab:ta3dgs_new_soccerfield}
\end{table*}

\begin{table*}[h]
\centering
\resizebox{\textwidth}{!}{
\begin{tabular}{c|ccc|ccc|ccc}
\toprule
\multirow{2}{*}{Frame} & \multicolumn{3}{c|}{\textbf{S1} $(\textrm{Warm} + \textrm{Densify})$} & \multicolumn{3}{c|}{\textbf{S2} $(\textrm{Warm} + \textrm{NoDensify})$} & \multicolumn{3}{c}{\textbf{S3} (GT Init)} \\
\cmidrule(lr){2-4} \cmidrule(lr){5-7} \cmidrule(lr){8-10}
 & Train (sec) & Splats & Size (MB) & Train (sec) & Splats & Size (MB) & Train (sec) & Splats & Size (MB) \\
\midrule
1 & 134.73 & 188,393 & 44.56 & 202.02 & 100,000 & \textbf{23.65} & 209.73 & 187,982 & 44.46 \\
2 & 164.55 & 355,786 & 84.15 & 197.19 & 100,000 & \textbf{23.65} & 208.50 & 189,927 & 44.92 \\
3 & 195.33 & 524,774 & 124.12 & 195.77 & 100,000 & \textbf{23.65} & 210.12 & 189,107 & 44.73 \\
4 & 225.10 & 691,359 & 163.52 & 194.16 & 100,000 & \textbf{23.65} & 212.10 & 188,876 & 44.67 \\
5 & 253.52 & 846,094 & 200.11 & 194.29 & 100,000 & \textbf{23.65} & 210.75 & 189,723 & 44.87 \\
\bottomrule
\end{tabular}
}
\caption{Warm start initialization analysis under different experimental setting. We compare \textbf{S1} $(\textrm{Warm} + \textrm{Densify})$, \textbf{S2} $(\textrm{Warm} + \textrm{NoDensify})$, and \textbf{S3} (GT Init) for the dynamic part of the scene over the first 5 time instance. The quantitative results clearly support our approach (\textbf{S2}).}
\label{tab:efficiency_growth}
\end{table*}

\subsubsection{Evaluation and Results}
We evaluate the rendering quality of approaches using the popular PSNR and LPIPS metric. We measure the rendering efficiency in frames-per-second (FPS). We assume synchronized well-calibrated camera intrinsic and extrinsic setting. For the first time step, we initialize the 3DGS model using a ground-truth point cloud exported from our Blender-based data generation plugin. For each subsequent time step, we initialize the optimization using the optimized 3DGS model from the immediately preceding frame. We disable the densification for the subsequent time step to reduce computational overhead, while adhering to warm-start for preserving temporal consistency.

\begin{figure}[h]
    \centering
\includegraphics[width=\linewidth]{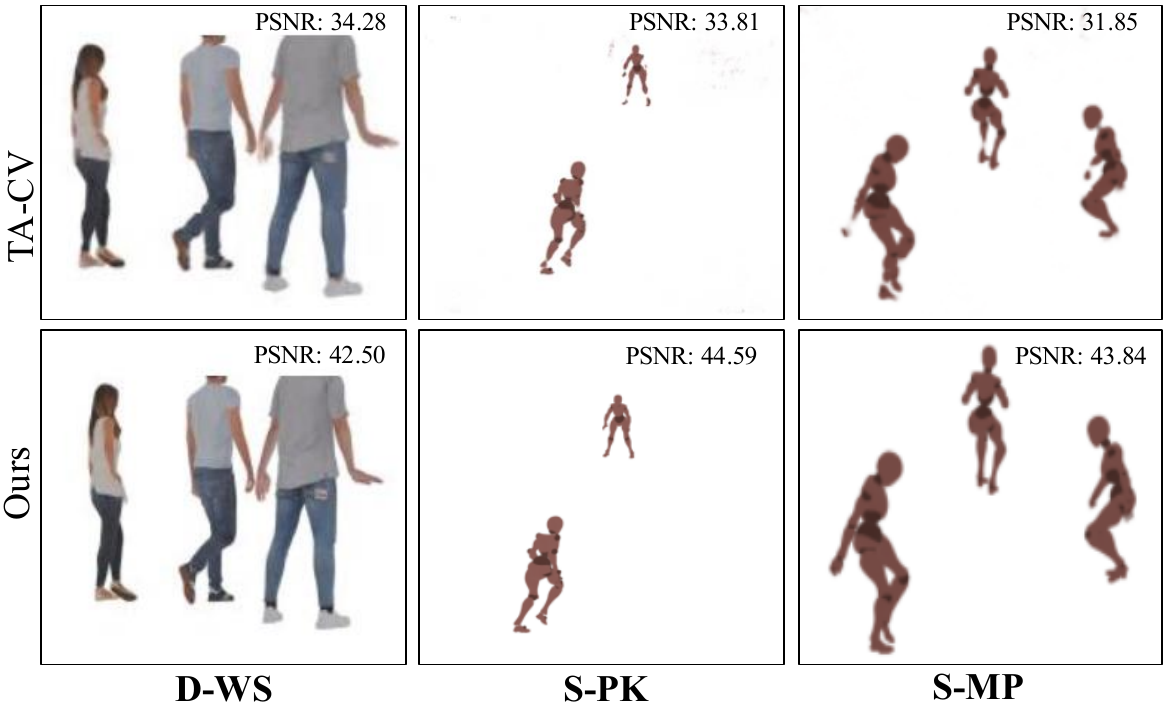}
\caption{Qualitative comparison with recent TACV \cite{zhang2026timearchivalcameravirtualizationsports} on D-WS, S-PK, and S-MP. Clearly, our approach shows better  results.}\label{fig:synthetic-dataset-results-1}
\end{figure}

\smallskip
\formattedparagraph{\textit{(i)} Results on D-WS, S-PK, and S-MP dataset.} For benchmarking, we compare our method against popular dynamic scene NVS methods such as D-NeRF \cite{pumarola2021d}, D-3DGS \cite{luiten2024dynamic}, 4DGS \cite{wu20244d}, and ST-GS \cite{li2024spacetime}, and others. For the first time step, we initialize all the 3DGS model with same $100,000$ points. The initial time frames and all subsequent time frames are trained for $8,000$ iterations. Such a take enables an efficient temporally chained optimization process while reusing geometric priors from earlier frames. Table \ref{tab:comparison_result_with_both_3dgs_and_nerf_approaches} provides the quantitative comparison results, demonstrating the benefits of our approach in rendering quality while managing favorable memory footprint and train time. Figure \ref{fig:synthetic-dataset-results-1} show comparison results with recent work \cite{zhang2026timearchivalcameravirtualizationsports}.

\smallskip
\formattedparagraph{\textit{(ii)} Results on Weather-Condition Soccerfield dataset.}
Since this dataset contains background with significant information, we decompose the scene into two components: (1) a static component $\textbf{A}$ and (2) the dynamic components $\textbf{B}$. $\textbf{A}$  corresponds to the stadium and background, while $\textbf{B}$ comprises the moving players. By separating the scene into a static component $\textbf{A}$  and a dynamic component $\textbf{B}$, we avoid repeatedly optimizing the full scene at every time step. The static background $\textbf{A}$ is reconstructed once, while $\textbf{B}$ focuses on the time-varying foreground modeling. This significantly reduces the model size for per-frame optimization and improves training efficiency. At rendering time, the two components are merged to produce the final scene.

To isolate the dynamic component, we construct residual masks for each camera view by comparing the ground-truth image with an $\textbf{A}$-only rendering, followed by RGB differencing, thresholding, and simple morphological refinement. The Gaussians in $\textbf{B}$ are then projected across all views, and a multi-view voting rule is applied to retain only Gaussians consistently supported by the residual masks. This helps suppress background leakage and preserve motion-consistent dynamic points. We also optionally apply a 3D Axis-Aligned Bounding Box (AABB)-based boundary rule to smooth the transition near the merge region. The final frame-wise scene is obtained by merging the Gaussians from the static component $\textbf{A}$ with the retained Gaussians from $\textbf{B}$ into a single 3DGS representation, which is then used for rendering. In practice, this $\textbf{A/B}$ separation performs best when both components share identical camera parameters and similar lighting conditions; otherwise, the residual masks may become noisy.

For the first time step, we initialize the model with $100,000$ points, and trained for $30000$ and $5200$ for \textbf{A} and \textbf{B}, respectively, Table \ref{tab:ta3dgs_new_soccerfield} shows the quantitative results obtained using our approach. It is observed that by using Warm-start approach, we can deploy dynamic NVS solution for large-scale dynamic novel synthesis that is memory efficient with excellent image synthesis quality, and thus suitable for retrospective archival and replay.

\subsection{Ablations}
We conducted all our ablations on the proposed Snow SoccerField dataset. To show the practical importance of our approach, we performed this ablation under three setting:

\begin{itemize}
    \item \textbf{S1} ($\textrm{Warm} + \textrm{Densify}$): start the experiment with warm-start initialization and enable 3DGS densification.
    \item \textbf{S2} ($\textrm{Warm} + \textrm{NoDensify}$): start the experiment with warm-start initialization but disable 3DGS densification.
    \item \textbf{S3} (GT Init): for every time step, initialize the splats using GT point cloud.
\end{itemize}

\smallskip
\formattedparagraph{\textit{(i)} Warm-Start Initialization Analysis.} 
The objective of this ablation is to evaluate whether warm-start initialization without densification training across sequential frames is scalable and efficient take to the problem or not. 

From statistical result presented in Table \ref{tab:efficiency_growth}, it is evident that \textbf{S1} suffers from a severe scalability problem. Here, enabling densification leads to continuous accumulation of Gaussians splats for the dynamic players over time, thus rapidly inflating the model capacity, i.e., growing from $\sim$$188$K to $\sim$$850$K splats over just 5 frames. This leading to a $5\times$ increase in its memory footprint. Meanwhile, \textbf{S3} provides a impractical setting yet a stable reference. This experimental setting resets the method with new splats via GT point cloud initialization for each time instance. While the results could be impressive, it is practically not feasible due to high-compute requirement either via deployment of LiDAR sensors or running COLMAP for each time instance, and therefore serves as a stable baseline. On the contrary, our approach i.e., \textbf{S2} maintains a strictly fixed capacity (e.g., 100K Splats), avoiding temporal splat accumulation while keeping both runtime and storage predictable. Overall, this ablation clearly highlight the scalability limitations of unconstrained densification for sequential warm-start training and suitability of our approach in efficient retrospective NVS for long dynamic scene video.

\smallskip
\formattedparagraph{\textit{(ii)} Image Synthesis Quality and Practical Feasibility Analysis.} We further analyzed the trade-off between initialization efficiency and the final merged scene rendering quality (measured in PSNR and LPIPS). Table \ref{tab:quality_comparison} provides the quantitative results across 15 frames of the Snow SoccerField sequence. While \textbf{S1} achieves consistently better PSNR and LPIPS metrics, this empirical improvement is entirely driven by its continuously expanding number of splats and model capacity, which we have shown to be highly impractical for longer sequences due to excessive computational overhead (refer to Table \ref{tab:efficiency_growth}). On the contrary, \textbf{S2} exhibits a mild performance improvement over time, offering a better trade-off between rendering quality, efficiency, and model capacity for the practical deployment of retrospective dynamic scene NVS. 

\subsection{Limitations}

\smallskip
\formattedparagraph{Capture assumptions.} Our formulation assumes (i) accurate camera calibration and (ii) temporal synchronization across views. In real deployments, small errors in intrinsics, extrinsics, rolling-shutter effects, or sub-frame asynchronization can lead to multi-view inconsistencies that may degrade reconstruction quality and, this can lead to accumulation of error over time due to warm-chain setting. Moreover, our method is tailored to static camera rigs and extensions to moving or zooming cameras, or to continuously changing intrinsics are not directly addressed.

\formattedparagraph{Initialization dependence.} We initialize the first frame using point cloud, after which optimization proceeds via warm-start. If the initial point cloud is too sparse or biased (e.g., due to reflections, textureless surfaces, or a limited baseline), the fixed-budget warm chain may not fully recover missing geometry. Improving robustness under imperfect initialization is an important direction.

\begin{table}[h]
\centering
\resizebox{\linewidth}{!}{
\begin{tabular}{c|cc|cc}
\toprule
\multirow{2}{*}{Frame} & \multicolumn{2}{c|}{ \textbf{S1} $(\textrm{Warm} + \textrm{Densify})$} & \multicolumn{2}{c}{\textbf{S2} $(\textrm{Warm} + \textrm{NoDensify})$} \\
\cmidrule(lr){2-3} \cmidrule(lr){4-5} 
 & PSNR & LPIPS & PSNR & LPIPS \\
\midrule
1  & 28.39 & 0.395 & 27.23 & 0.426 \\
2  & 29.12 & 0.373 & 27.73 & 0.412 \\
3  & 29.87 & 0.353 & 27.98 & 0.406 \\
4  & 30.60 & 0.335 & 28.10 & 0.402 \\
5  & 31.25 & 0.325 & 28.22 & 0.400 \\
\bottomrule
\end{tabular}
}
\caption{Quality Comparison (PSNR $\uparrow$ and LPIPS $\downarrow$) across 5 frames from the Snow SoccerField dataset under \textbf{S1} and \textbf{S2} ablation settings. Metrics are computed on the final merged renderings.}
\label{tab:quality_comparison}
\end{table}

\section{Conclusion}
\label{sec:conclusion}
In this paper, we studied retrospective NVS of dynamic scenes under synchronized multi-view capture, and revisited the necessity of explicit temporal deformation constraints for dynamic 3D Gaussian Splatting. Our central observation is that, in a well calibrated multi-camera rigs, each time instant is already strongly constrained by multi-view geometry. Building on this premise, we introduced a warm-start, densification-free 3DGS pipeline that propagates optimized Gaussians over time while keeping model size and compute predictable. Across the benchmark datasets, the proposed approach achieves high-fidelity retrospective image rendering quality with excellent practical scalability, showing that strong multi-view constraints can reduce reliance on complex temporal coupling. To complement our approach, we proposed a standardized, dynamic, multi-view dataset generation API and benchmarking framework in Blender to reduce friction caused by inconsistent coordinate conventions and dataset formats. Together, our contributions aim to improve both the practical efficiency of time-archival dynamic NVS and the reproducibility of comparisons across dynamic NeRF and 3DGS methods.

\section*{Acknowledgment} The authors thank High Performance Research Computing (HPRC) at Texas A\&M University, College Station Texas, USA for providing us with the startup credits for utilizing the GPU-server facility.

{
    \small
    \bibliographystyle{ieeenat_fullname}
    \bibliography{main}
}

\maketitlesupplementary

    

\section{More Detail on Blender API}
While Neural Radiance Fields (NeRF) and 3D Gaussian Splatting (3DGS) have achieved remarkable progress in novel view synthesis 
, preparing multi-view datasets across different methods remains a bottleneck due to varying format requirements and coordinate system conventions 
To alleviate the extensive manual setup and debugging typically required for dataset conversion, we introduce our comprehensive Blender add-on designed to streamline the generation of synthetic datasets.

It facilitates the extraction of consistent camera parameters and rendered images over time, bridging the gap between synthetic scene creation in Blender and ready-to-train pipelines for various 3DGS and NeRF variants in just a few clicks 
It is particularly well-suited for dynamic scene evaluation and complex camera virtualization tasks, such as those involving sports broadcasting 

\subsection{Key Features}
Our proposed tool significantly accelerates the data preparation and empirical evaluation processes through the following core functionalities:

\begin{itemize}
    \item \textbf{Versatile Camera Generation \& Stadium Layout:} The add-on supports standard sampling layouts, including upper hemisphere, full sphere, and elliptical rings 
    Crucially for large-scale and sports-focused applications, it introduces a specialized \textit{Stadium Layout} designed for soccer and football stadiums 
    This feature allows users to define field dimensions, generate grid-based camera tiles 
    , and automatically position goal-line cameras 
    , ensuring comprehensive multi-view coverage for methods like TACV 
    
    \item \textbf{Flexible Dataset Export Modes:} To support modern point-based rendering techniques, RS Studio exports point clouds using either a 3DGS COLMAP surface mode (sampled directly from the mesh surface) or a 3DGS depth mode (generated via ray intersections from RS Studio cameras) 
    
    \item \textbf{Dynamic Camera Tracking:} For downstream evaluation and instant render testing of dynamic models 
    , the tool exports per-frame camera tracks with intrinsics and extrinsics formatted as JSON files 
    Supported tracking modes include keyframed animation, automated orbit paths, linear interpolation, and target following 
    
    \item \textbf{Dataset Import and Pose Validation:} Users can import existing datasets in standard NeRF synthetic or custom TACV formats 
    to visually verify camera poses within the 3D environment or reuse existing configurations for new data generation 
    
    \item \textbf{Geographically-Aware Lighting:} Through integration with physical solar models, the tool provides adjustable indoor and outdoor lighting driven by specific geographic coordinates (latitude and longitude) and time 
    , enabling robust illumination augmentation for synthetic scenes.
\end{itemize}

By open-sourcing RS Studio, we aim to provide the community with a robust pipeline that lowers the barrier to entry for constructing complex, temporal, and large-scale synthetic datasets.


\section{Additional Ablation Studies}
\label{sec:supp_ablation}
Here, we provide additional analyses supporting the two central claims of the paper: (i) our time-archival 3DGS formulation preserves the practical efficiency of Gaussian splatting for retrospective rendering, and (ii) warm-start training primarily exploits local temporal smoothness rather than any implicit temporal coupling in the objective. Unless otherwise stated, all evaluations follow the same camera calibration, train/val/test split protocol, and training implementation described in the main paper.

\begin{table}[h]
\centering
\caption{Real-time rendering comparison. TA-3DGS preserves the native rendering efficiency of Gaussian splatting while operating at full HD resolution (1920$\times$1080). On a single A40 GPU, our method achieves real-time rendering across three TACV datasets, reaching 65.8 FPS on DWS, 55.9 FPS on SPK, and 59.9 FPS on SMP. Prior dynamic Gaussian methods such as 4DGS and FreeTimeGS also report real-time rendering, although the reported numbers are measured on different GPUs and resolutions.}
\label{tab:fps_compare_realtime}
\resizebox{\linewidth}{!}{
\begin{tabular}{l c c c}
\toprule
Method & GPU & Rendering Resolution & FPS $\uparrow$ \\
\midrule
4DGS \cite{wu20244d} & RTX 3090 & 800$\times$800 & 82 \\
FreeTimeGS \cite{wang2025freetimegs} & RTX 4090 & 1920$\times$1080 & 450 \\
\midrule
TA-3DGS (ours, DWS) & A40 & 1920$\times$1080 & 65.8 \\
TA-3DGS (ours, SPK) & A40 & 1920$\times$1080 & 55.9 \\
TA-3DGS (ours, SMP) & A40 & 1920$\times$1080 & 59.9 \\
\bottomrule
\end{tabular}
}
\end{table}

\subsection{Real-Time Rendering Throughput at Full HD}
\label{sec:supp_realtime}

Table~\ref{tab:fps_compare_realtime} reports rendering throughput at full HD (1920$\times$1080) on a single NVIDIA A40 GPU. TA-3DGS achieves real-time performance across all three TACV sequences, reaching 65.8 FPS on DWS, 55.9 FPS on SPK, and 59.9 FPS on SMP. These results indicate that our time-archival representation retains the native runtime advantages of Gaussian splatting despite operating at high resolution.

\paragraph{Resolution-normalized throughput.}
Since prior dynamic Gaussian approaches report FPS under different resolutions and GPU configurations, direct FPS comparisons are not strictly controlled. For additional context, we report a resolution-normalized throughput in megapixels per second (MPix/s), computed as $\mathrm{MPix/s}=\mathrm{FPS} \cdot H\cdot W/10^6$. Under this metric, TA-3DGS achieves 136.4 MPix/s (DWS), 115.9 MPix/s (SPK), and 124.2 MPix/s (SMP) at $1920\times1080$. By comparison, 4DGS reports 52.5 MPix/s at $800 \times 800$, while FreeTimeGS \cite{wang2025freetimegs} reports substantially higher throughput at $1920 \times 1080$ on an RTX 4090. We emphasize that these numbers remain indicative only, as GPU architecture and implementation details differ across works.

\subsection{Warm-Start Test}
\label{sec:supp_distant_warmstart}

A key interpretability question is whether warm-starting yields benefits because the optimization implicitly encodes a temporal constraint (e.g., through hidden coupling in the loss), or simply because adjacent frames are typically locally smooth. To isolate this effect, we introduce an \textbf{``distant warm-start''} baseline on DWS (Table~\ref{tab:dws_distant_warm_baseline}). This baseline keeps the dataset, loss, and training pipeline fixed, but replaces the standard adjacent-frame initialization with an initialization from a temporally distant checkpoint; the result is reported on frames 40--49.

The distant warm-start baseline exhibits a \textbf{``large quality degradation''}, dropping PSNR from 42.50 to 26.80 (absolute drop: 15.70 dB; relative: $-36.94\%$) and increasing LPIPS from 0.0111 to 0.0246 (absolute increase: 0.0135; relative: $+121.99\%$). This controlled perturbation supports the interpretation that warm-start primarily leverages local temporal continuity and a favorable basin of attraction for optimization, rather than relying on any implicit temporal coupling embedded in the objective formulation.

\subsection{Extended Warm-Start Analysis}
\label{sec:supp_warmstart_extended}

The main paper reports warm-start ablations over a short temporal window i.e., 5 frame window. Here, we extend the evaluation to 15 time instances for the dynamic component (Table~\ref{tab:warm_start_extended}), comparing:
\textbf{S1} (Warm + Densify), \textbf{S2} (Warm + NoDensify, ours), and \textbf{S3} (GT Init per frame).
This extended analysis clarifies two practical properties: quality robustness over longer warm-start chains and the effect of densification on apparent quality gains.

\paragraph{Quality trends.}
Across 15 frames, S1 achieves consistently higher PSNR and lower LPIPS than S2 and S3; however, this improvement is expected because densification increases model capacity over time, which is precisely the scalability issue highlighted in the main paper. In contrast, S2 maintains a fixed Gaussian budget and therefore prioritizes stability in representation size and speed. Despite this constraint, S2 remains competitive with the per-frame GT initialization baseline S3. Averaged across the 15 frames, S2 attains 28.21 dB PSNR and 0.3998 LPIPS, while S3 attains 28.37 dB PSNR and 0.3945 LPIPS. The small gap indicates that warm-starting with fixed capacity is able to preserve most of the per-frame reinitialization quality while avoiding its computational overhead and impractical sensing requirements.

\paragraph{Interpretation.}
The extended results reinforce the paper's central claim: under synchronized multi-view constraints, a temporally chained optimization strategy can remain stable over time even without explicit deformation constraints. Meanwhile, the superior quality of S1 should be interpreted in light of its expanding model complexity (enabled by densification), rather than as evidence that explicit temporal coupling is necessary.

\subsection{Per-Frame Training Throughput Under Fixed-Capacity Warm-Start}
\label{sec:supp_speed_extended}

Table~\ref{tab:speed_extended} reports the per-iteration time for S2 across the same 15-frame window. Even with a fixed number of Gaussians, the iteration time increases from 12.34 ms (frame 1) to 34.50 ms (frame 15), with an average of 24.91 ms per iteration. This increase is consistent with the fact that the rasterization workload of Gaussian splatting depends not only on the \emph{count} of Gaussians but also on scene-dependent visibility patterns and screen-space coverage, which can change over time for dynamic sequences.

\paragraph{Practical implication.}
Despite this variability, S2 maintains predictable memory and representation size (no densification) while sustaining practical training throughput across time. This observation complements the real-time rendering results (Section~\ref{sec:supp_realtime}) and supports the overall deployment motivation of time-archival retrospective rendering in long dynamic events.

\begin{table}[t]
\centering
\caption{\textbf{Effect of abnormal distant warm-start on DWS.}
Compared with the standard warm-start setting, initializing a target segment from a temporally distant checkpoint causes a substantial quality drop on DWS. This supports the interpretation that warm-start primarily benefits from \emph{local temporal smoothness}, rather than any hidden temporal coupling in the loss design.
}
\label{tab:dws_distant_warm_baseline}
\resizebox{0.72\linewidth}{!}{
\begin{tabular}{lcc}
\toprule
\textbf{Setting} & \textbf{PSNR}$\uparrow$ & \textbf{LPIPS}$\downarrow$ \\
\midrule
Standard warm-start & 42.5027 & 0.0111 \\
Distant warm-start baseline & 26.8031 & 0.0246 \\
\midrule
Absolute drop / increase & -15.6996 & +0.0135 \\
Relative change & -36.94\% & +121.99\% \\
\bottomrule
\end{tabular}
}
\vspace{2mm}
\footnotesize

\textit{Note:} The distant warm-start baseline keeps the same dataset and training pipeline, but replaces normal adjacent-frame initialization with an abnormal initialization from a temporally distant checkpoint. The reported distant warm-start result is measured on frames 40--49 of DWS.
\end{table}
\begin{table}[t]
\centering
\caption{\textbf{Extended Analysis of Warm Start Initializations.} We provide a detailed comparison of S1 (Warm+Densify), S2 (Warm+NoDensify), and S3 (GT Init) across 15 time instances. These extended results for the dynamic part of the scene further support the robustness of our approach (S2) beyond the 5 frames presented in the main paper.}
\label{tab:warm_start_extended}
\resizebox{\columnwidth}{!}{%
\begin{tabular}{@{}c|ccc|ccc@{}}
\toprule
& \multicolumn{3}{c|}{PSNR $\uparrow$} & \multicolumn{3}{c}{LPIPS $\downarrow$} \\
Frame & S1 & S2 & S3 & S1 & S2 & S3 \\ \midrule
1  & 28.386 & 27.229 & 28.356 & 0.3948 & 0.4257 & 0.3954 \\
2  & 29.117 & 27.727 & 28.394 & 0.3733 & 0.4118 & 0.3942 \\
3  & 29.867 & 27.980 & 28.343 & 0.3532 & 0.4058 & 0.3948 \\
4  & 30.595 & 28.103 & 28.345 & 0.3350 & 0.4024 & 0.3952 \\
5  & 31.253 & 28.221 & 28.367 & 0.3246 & 0.3998 & 0.3951 \\
6  & 31.856 & 28.255 & 28.320 & 0.3163 & 0.3982 & 0.3943 \\
7  & 32.222 & 28.270 & 28.351 & 0.3108 & 0.3972 & 0.3943 \\
8  & 32.643 & 28.317 & 28.345 & 0.3071 & 0.3959 & 0.3942 \\
9  & 32.839 & 28.375 & 28.388 & 0.3035 & 0.3953 & 0.3940 \\
10 & 33.235 & 28.381 & 28.342 & 0.3005 & 0.3948 & 0.3946 \\
11 & 33.356 & 28.424 & 28.439 & 0.2990 & 0.3947 & 0.3948 \\
12 & 33.519 & 28.412 & 28.388 & 0.2975 & 0.3941 & 0.3949 \\
13 & 33.707 & 28.431 & 28.442 & 0.2964 & 0.3944 & 0.3940 \\
14 & 33.773 & 28.491 & 28.421 & 0.2950 & 0.3938 & 0.3940 \\
15 & 33.818 & 28.484 & 28.361 & 0.2944 & 0.3934 & 0.3938 \\ \bottomrule
\end{tabular}
}
\end{table}

\begin{table}[h]
\centering
\caption{\textbf{Per-frame Training Speed of S2.} Detailed iteration time (ms) for our proposed S2 setting across the extended sequence.}
\label{tab:speed_extended}
\begin{tabular}{@{}lc@{}}
\toprule
Frame & Iteration Time (ms) \\ \midrule
1  & 12.340 \\
2  & 14.581 \\
3  & 16.810 \\
4  & 18.900 \\
5  & 20.887 \\
6  & 22.770 \\
7  & 24.506 \\
8  & 25.809 \\
9  & 27.194 \\
10 & 28.618 \\
11 & 29.888 \\
12 & 31.100 \\
13 & 32.348 \\
14 & 33.395 \\
15 & 34.498 \\ \bottomrule
\end{tabular}
\end{table}

\end{document}